\documentclass[10pt,twocolumn,letterpaper]{article}
\usepackage[pagenumbers]{wacv}
\usepackage{multirow}
\usepackage{makecell}
\usepackage{tikz}
\usetikzlibrary{arrows.meta,positioning,calc}
\def\Approach{TokenStream\xspace}
\newcommand{\MultiDB}{Multi-DB\xspace}
\newcommand{\myheading}[1]{\vspace{1mm}\noindent{\textbf{#1}}}

\usepackage[accsupp]{axessibility}
\definecolor{wacvblue}{rgb}{0.21,0.49,0.74}
\usepackage[breaklinks,colorlinks,allcolors=wacvblue]{hyperref}
\hypersetup{pdftitle={TokenStream: Multi-Subject Personalized Video Generation with Token-wise LoRA and Guided Attention},pdfauthor={Hung Nguyen, Quang Qui-Vinh Nguyen, Khoi Nguyen}}
\def\wacvPaperID{1317}
\def\confName{WACV}
\def\confYear{2027}
\title{\Approach: Multi-Subject Personalized Video Generation with Token-wise LoRA and Guided Attention}
\author{Hung Nguyen$^{1}$\footnotemark[1]\thanks{Work done while at Qualcomm.} \qquad Quang Qui-Vinh Nguyen$^{2}$ \qquad Khoi Nguyen$^{2}$\\
$^{1}$National University of Singapore, Singapore\\
$^{2}$Qualcomm AI Research\thanks{Qualcomm AI Research is an initiative of Qualcomm Technologies, Inc.}\\
{\tt\small hungnm@comp.nus.edu.sg \quad quanguy@qti.qualcomm.com \quad khoi@qti.qualcomm.com}
}

\begin{document}
\maketitle
\begin{abstract}
We introduce \textbf{\Approach}, an optimization-based framework that composes independently learned subject identities in text-to-video generation without a reference-video training dataset or joint optimization of each new subject combination.
Learning from a single reference image makes subject appearance difficult to separate from static-image bias, while merging independently trained adapters can mix identities.
\textbf{Token-wise LoRA} (ToRA) restricts low-rank updates to the cross-attention value projections of subject-associated text tokens, providing a separate adaptation path for each subject.
A \textbf{T2V-guided stream} copies queries and keys from the frozen base model during early denoising to reuse its layout and motion priors while retaining personalized values.
We also introduce \textbf{\MultiDB}, a 30-subject, 122-prompt benchmark with appearance-free prompts and varied interactions.
On \MultiDB at 1.3B, \Approach improves identity preservation from 0.422 for DreamBooth to 0.467 and textual alignment from 0.652 to 0.688.
\end{abstract}

\begin{figure*}[t]
    \centering
    \includegraphics{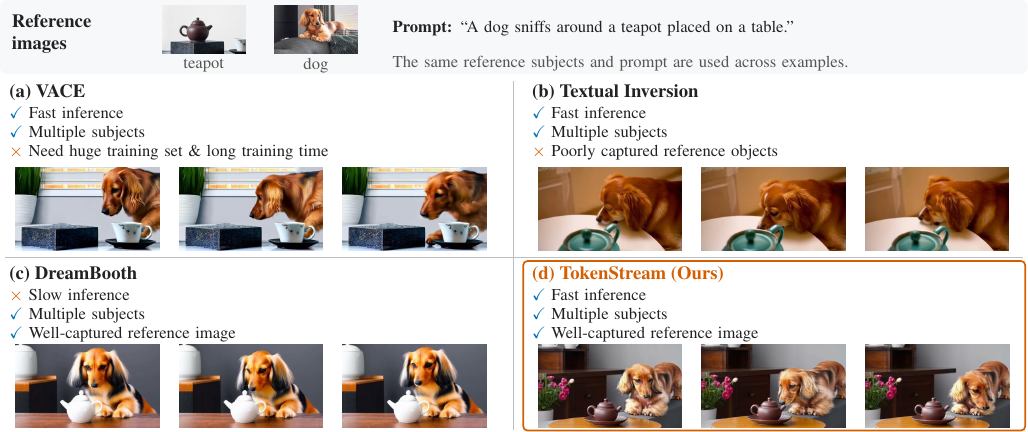}
    \caption{\textbf{Multi-subject video personalization with \Approach.}
    \Approach uses a single reference image per subject and composes independently trained subject adapters at inference time.
    Compared with Textual Inversion and DreamBooth-style adaptation, it better preserves subject appearance while retaining the motion and prompt-following behavior of the pretrained T2V model.}
    \label{fig:teaser}
\end{figure*}
\section{Introduction}
\label{sec:intro}

\textbf{Multi-subject personalized video generation} aims to generate a video that follows a text prompt while faithfully depicting multiple target subjects given as reference images, with applications from filmmaking to personalized advertising.
The task is difficult because it requires an accurate correspondence between visual subjects and their textual descriptions, and consistent identity, motion and interaction across frames.
Models adapted to static subject images also struggle to animate them, producing rigid or stationary appearances.

Existing methods fall into two categories.
\textit{Training-based approaches}~\cite{liu2025phantom,jiang2025vace,fei2025skyreels,deng2025magref,hu2025hunyuancustom,xu2025compositional,zhang2025kaleido,li2025bindweave,zhang2025tora2,cai2025omnivcus,hu2025polyvivid,zhong2025concat,deng2025cinema,liang2025movie,chen2024videoalchemy} inject reference subjects during model training.
They are effective but require large-scale, high-quality video-reference pairs that are expensive to collect, and fine-tuning on such data can degrade the generative quality of the pretrained T2V model.
\textit{Optimization-based approaches}~\cite{chen2023videodreamer,wang2024customvideo,jiang2024videobooth,wei2024dreamvideo,chen2024disenstudio,wu2024motionbooth,huang2025videomage,abdal2025dynamic} instead optimize subject-specific parameters, and most derive from DreamBooth~\cite{ruiz2023dreambooth} or Textual Inversion (TI)~\cite{gal2022image}.
DreamBooth improves appearance fidelity by updating weights or adapters, but multi-subject composition is hard because independently trained adapters interfere.
TI learns one token embedding per subject and composes naturally, yet its capacity is limited to the text embedding space and it captures incomplete identity details.
Both also overfit static reference images, weakening motion dynamics and prompt adherence.

We therefore introduce \Approach, with two components.
\textit{(1) Token-wise LoRA (ToRA).}
Unlike TI, which learns a single embedding vector per subject token, ToRA assigns a dedicated LoRA~\cite{hu2022lora} to each subject-relevant token within the Text-to-Value projection of every cross-attention layer.
This greatly expands capacity for fine-grained identity while keeping the compositional flexibility of text-driven control, so multiple subjects combine without additional optimization.
\textit{(2) T2V-guided stream.}
At inference, an auxiliary stream driven by the frozen T2V model steers the denoising trajectory of the personalized stream, reusing the pretrained motion and layout priors while retaining the subject-specific value features learned by ToRA.
Unlike image-based methods that transfer $K$ and $V$, we copy $Q$ and $K$ to transfer attention structure while letting the main stream inject personalized appearance.

The current mainstream benchmark, OpenS2V~\cite{yuan2025opens2v}, exhibits limitations: (i) its prompts contain limited motion diversity and subject interaction, and (ii) detailed appearance descriptions undermine the personalization objective, as the model should learn subjects from reference images rather than the prompt itself.
To enable more realistic evaluation, we introduce \MultiDB, a new benchmark built on DreamBench~\cite{ruiz2023dreambooth}.
\MultiDB contains 30 subjects and 122 prompts, including 76 two-subject and 46 three-subject compositions, with richer motion dynamics and higher compositional complexity.
The prompts are designed to avoid explicit appearance descriptions.
Thus, \MultiDB is designed to test three specific weaknesses of existing benchmarks: prompts that reveal the subject appearance, simple subject interactions, and limited motion diversity.

We build our proposed method on Wan2.1~\cite{wan2025wan}, a pretrained video generation framework, and evaluate it on both our proposed benchmark and OpenS2V~\cite{yuan2025opens2v}.
We compare our approach with TI and DreamBooth as optimization-based baselines, as well as training-based methods including VACE~\cite{jiang2025vace}, Phantom~\cite{liu2025phantom}, MAGREF~\cite{deng2025magref}, and Skyreels-A2~\cite{fei2025skyreels}.
Our method improves identity preservation and textual alignment over the evaluated TI and DreamBooth baselines, while offering a complementary trade-off to training-based approaches without large-scale reference-video training.

In summary, our main contributions are as follows:
\begin{itemize}
    \item We present \textbf{\Approach}, a novel optimization-based framework for multi-subject personalized video generation, featuring ToRA, a token-wise LoRA fine-tuning technique for precise textual control.
    \item To address subject overfitting that limits motion and prompt following, we introduce the T2V-guided stream, an inference-time method that transfers attention structure while retaining personalized values.
    \item We establish \textbf{\MultiDB}, a new benchmark for multi-subject video personalization designed to reduce prompt leakage and increase motion and interaction complexity.
\end{itemize}

\section{Related Work}
\label{sec:related_work}

\myheading{Video Generation Model.}
Recent progress in diffusion-based video generation has markedly expanded creative capabilities.
Early video diffusion models relied on U-Net denoisers, as in Stable Video Diffusion~\cite{blattmann2023stable} and AnimateDiff~\cite{guo2023animatediff}, while recent systems increasingly adopt DiT-based architectures~\cite{peebles2023scalable}.
These models mainly follow two designs: MMDiT-based models, such as CogVideoX~\cite{yang2024cogvideox} and HunyuanVideo~\cite{kong2024hunyuanvideo}, and cross-attention-based models, such as Wan2.1~\cite{wan2025wan}, Cosmos~\cite{ali2025world}, and LTX-Video~\cite{hacohen2024ltx}.
MMDiT jointly attends over text and visual tokens, making precise token-level intervention less direct.
In contrast, cross-attention-based architectures decouple self-attention from cross-attention, enabling targeted modulation of subject-relevant text tokens.
Capitalizing on this advantage, we adopt Wan2.1 as our base T2V model.

\myheading{Personalized/Subject-driven Image Generation.}
Personalization methods adapt pretrained generative models to learn a subject from one or a few reference images.
Training-based methods~\cite{li2023blip,ye2023ip,Zhang_2024_CVPR,Wei2023ELITEEV,he2025disenvisioner} introduce subject information through learned encoders or adapters, but require large-scale, high-quality image-reference pairs.
Optimization-based methods update text embeddings~\cite{gal2022image,alaluf2023neural,voynov2023p+,nguyen2025csd} or model weights~\cite{ruiz2023dreambooth,gal2023designing}.
Textual Inversion (TI)~\cite{gal2022image} enables flexible subject composition via learned tokens but is limited by the text embedding space.
DreamBooth-style approaches better preserve appearance through weight updates, yet independently trained LoRAs do not compose well for multiple subjects.
Recent extensions such as ConceptSplit~\cite{lim2025conceptsplit} and TokenVerse~\cite{garibi2025tokenverse} improve token-level expressiveness, but they focus on image generation.
A related scene-aware and agentic line explores controllable visual creation through scene reasoning and tool use, such as JarvisArt~\cite{lin2025jarvisart} and IR3D-Bench~\cite{liu2025ir3d}. Since these works operate at the planning, tool-use, or evaluation level, they are complementary to our model-level focus on subject binding rather than direct baselines.
In contrast, we target multi-subject personalized video generation, where identity must remain stable while motion and prompt-conditioned interactions are preserved.

\myheading{Multi-subject Personalized Video Generation.}
Personalized video generation aims to produce prompt-aligned videos that faithfully depict specific subjects given their reference images.
Existing approaches largely fall into two categories.
\textit{Training-based approaches}~\cite{liu2025phantom,jiang2025vace,fei2025skyreels,deng2025magref,hu2025hunyuancustom,xu2025compositional,zhang2025kaleido,li2025bindweave,zhang2025tora2,cai2025omnivcus,hu2025polyvivid,zhong2025concat,deng2025cinema,liang2025movie,chen2024videoalchemy} incorporate subject information directly during model training.
For example, Phantom~\cite{liu2025phantom} fuses subject images with video and text modalities, and ID-Crafter~\cite{pan2025idcrafter} studies compositional multi-subject generation through VLM-grounded online reinforcement learning.
These methods are powerful but rely on large-scale video-reference pairs that are costly to curate and computationally intensive to train on.

\textit{Optimization-based approaches}~\cite{chen2023videodreamer,wang2024customvideo,jiang2024videobooth,wei2024dreamvideo,chen2024disenstudio,wu2024motionbooth,huang2025videomage,abdal2025dynamic} personalize generative models by directly optimizing subject-specific parameters.
Most build upon DreamBooth~\cite{ruiz2023dreambooth} or TI~\cite{gal2022image}.
VideoDreamer~\cite{chen2023videodreamer}, CustomVideo~\cite{wang2024customvideo}, DisenStudio~\cite{chen2024disenstudio}, and DynamicConcept~\cite{abdal2025dynamic} study subject appearance or motion adaptation.
VideoMage~\cite{huang2025videomage} learns subject and motion LoRAs using a motion-reference video and combines them through optimized fusion and collaborative sampling.
Our setting uses one image per subject and no motion-reference video; a controlled comparison with recent video-specific optimization methods remains future work.
Other works explicitly model multi-subject interactions~\cite{gu2023mix,kong2024omg,kumari2023multi}, yet still depend on extra fine-tuning for strong composition.
By contrast, our method supports flexible textual control and multi-subject video personalization without additional optimization steps.

\section{Method}
\label{sec:method}

\begin{figure*}[t]
    \centering
    \includegraphics{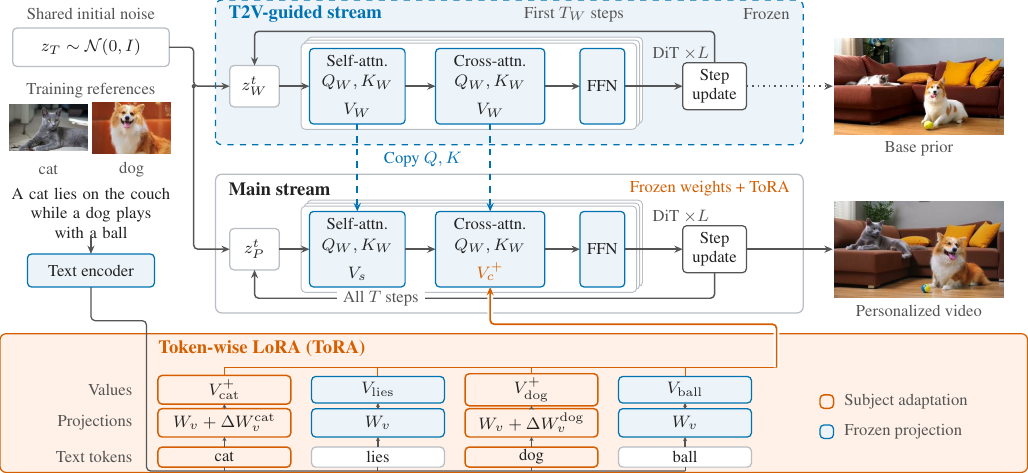}
    \caption{\textbf{Overall framework of \Approach.} Subject adapters are learned independently and applied only to their text-token spans in the value projection. The frozen T2V-guided stream transfers $Q,K$ attention features during the first $T_W$ denoising steps; the personalized stream retains its values and continues for all $T$ steps. Each model evaluation traverses $L$ self-attention--cross-attention--FFN blocks. Both streams share frozen weights but incur additional guidance computation.}
    \label{fig:framework}
\end{figure*}

\myheading{Problem statement.}
Given $N$ subjects, each represented by a \textbf{single image} $x_i$, and a user-provided text prompt $c$, our goal is to generate a video containing these $N$ subjects and complying with $c$.

\myheading{Method overview.}
Our \Approach is an optimization-based framework built on Wan2.1~\cite{wan2025wan}, a cross-attention-based T2V model.
It comprises Token-wise LoRA (ToRA) for subject adaptation during fine-tuning and a T2V-guided stream, activated only at inference time, as shown in \cref{fig:framework}.
We first introduce ToRA, then present the T2V-guided stream for mitigating static-image overfitting, and finally describe the \MultiDB benchmark.

\subsection{Token-wise LoRA}
\label{subsec:ToRA}

Textual Inversion (TI)~\cite{gal2022image} is a common approach for multi-subject personalized video generation.
It learns textual embeddings for subject-relevant tokens, making it flexible for composition, but its capacity is limited by the text embedding space.
DreamBooth~\cite{ruiz2023dreambooth} improves identity preservation by adapting model weights, yet independently trained LoRAs are difficult to combine for arbitrary multi-subject prompts.

To address this limitation, we propose Token-wise LoRA (ToRA), a LoRA~\cite{hu2022lora}-based fine-tuning strategy that operates on the cross-attention layers.
Given a prompt, each subject is associated with one or more subject-relevant tokens, typically the original noun phrase such as ``dog'' or ``toy car''.
We keep the text encoder and its vocabulary frozen; no new learned identifier embedding is introduced.
Routing assigns each subject a disjoint span of token indices, rather than attaching one adapter to every occurrence of the same word.
The corresponding token set is denoted as $\mathcal{S}$ for a single subject and $\mathcal{S}_k$ for subject $k$ in multi-subject generation.
For each cross-attention layer $l$, let $W_v^l$ denote the original Text-to-Value (To-V) projection.
Given subject-relevant token indices $\mathcal{S}$, we learn a low-rank update $\Delta W_v^l$ that is applied only to tokens in $\mathcal{S}$:
\begin{equation}
    V_i^l =
    \begin{cases}
        (W_v^l + \Delta W_v^l)e_i, & i \in \mathcal{S}, \\
        W_v^l e_i, & i \notin \mathcal{S},
    \end{cases}
    \label{eq:single_tora}
\end{equation}
where $e_i$ denotes the $i$-th text embedding and $\Delta W_v^l=B^l A^l$ is a rank-$r$ update.
The factors are shared across the tokens in one subject span within a layer, and learned separately for each subject and layer.

For multi-subject generation, each subject $k$ has its own ToRA weights $\Delta W_{v,k}^l$ and subject-token set $\mathcal{S}_k$.
At inference time, the value projection is computed as:
\begin{equation}
    V_i^l =
    \begin{cases}
        (W_v^l + \Delta W_{v,k}^l)e_i, & i \in \mathcal{S}_k,\; k \in \{1,\ldots,N\}, \\
        W_v^l e_i, & i \notin \bigcup_{k=1}^{N}\mathcal{S}_k.
    \end{cases}
    \label{eq:multi_tora}
\end{equation}
This formulation ensures that only the subject-specific tokens receive their corresponding adapted transformation, while all other tokens are processed using the original weights.
Consequently, multi-subject composition requires only loading the corresponding ToRAs, without weight averaging or joint re-optimization.
If a subject noun is split into multiple text tokens, the full span receives the same subject-specific ToRA.

\myheading{Training objective and reference-image handling.}
The reference image $x$ is encoded by the frozen Wan2.1 VAE into a \textbf{single-frame} latent $z_0=\mathcal{E}(x)$, i.e.\ the video length is set to one; the image is \emph{not} repeated across frames and no temporal mask is applied.
Given $\epsilon\sim\mathcal{N}(0,I)$, $t\sim\mathcal{U}(0,1)$ and the caption $c$ of the reference image, we form $z_t=(1-t)z_0+t\epsilon$ and minimize the conditional flow-matching loss
\begin{equation}
    \mathcal{L}
    = \mathbb{E}_{t,\epsilon}
    \left\lVert
    v_\theta\!\left(z_t, t, c\right) - \left(\epsilon - z_0\right)
    \right\rVert_2^2 ,
    \label{eq:flow_matching}
\end{equation}
where $v_\theta$ is the frozen Wan2.1 velocity predictor with the ToRA branch attached, with $\epsilon$ denoting Gaussian noise.
Only the low-rank matrices $\Delta W_{v}^{l}$ of the subject tokens receive gradients; the text encoder, VAE and base weights stay frozen.
Because the supervision is a single frame, ToRA sees no temporal signal, which motivates reusing a pretrained motion prior at inference rather than expecting the adaptation loss to learn motion.

Each ToRA is trained independently for a specific subject and stored for reuse.
In our implementation, learning one ToRA takes about 4 minutes and occupies roughly 5.6\,MB.
Once learned, these ToRAs can be loaded for new subject combinations without further fine-tuning; the quality limits of this composition are evaluated separately.

\myheading{Cost of token-wise binding.}
ToRA adds low-rank projections on selected text tokens without increasing the attention sequence length.
Adapter storage grows linearly with the subject count, and projection work also depends on the number of routed tokens.
In the measured two- to six-subject auxiliary runs, this overhead is small relative to video denoising: latency remains within 129--132\,s and peak allocated memory within 23.74--23.75\,GiB.

\subsection{T2V-Guided Stream}
\label{subsec:DSG}

Although ToRA expands the representational capacity for subject identity, learning from static reference images can entangle identity learning with motion modeling.
Unlike U-Net-based video models such as SVD~\cite{blattmann2023stable}, Wan2.1 adopts a unified DiT architecture without dedicated motion-specific modules.
Fine-tuning on still images can therefore encode static-image bias into the model, degrading motion fidelity and prompt alignment.

To resolve this issue, we propose a T2V-guided stream.
During inference, we retain the frozen base T2V model as a motion prior and copy its $Q$ and $K$ features into the personalized generation stream.
The guided stream shares the \emph{same weights} as the main stream and simply has no ToRA attached, so no second model is held in memory.
Both streams start from the same noise $z_T\sim\mathcal{N}(0,1)$.
At each timestep $t$ and layer $l$, we copy the query and key matrices, $Q_W^{t,l}$ and $K_W^{t,l}$, from the guided stream to the main stream, while the main stream computes its own value features.

For the self-attention layer, the output $z_s^{t,l}$ is computed as:
\begin{equation}
    z_s^{t,l} =
    \operatorname{Softmax}\left(\frac{Q_W^{t,l} K_W^{t,l\top}}{\sqrt{d}}\right)V_s^{t,l},
    \quad
    V_s^{t,l} = W_v^l z^{t,l-1}.
    \label{eq:self_guidance}
\end{equation}
For the cross-attention layer, the output $z_c^{t,l}$ is:
\begin{equation}
    z_c^{t,l} =
    \operatorname{Softmax}\left(\frac{Q_W^{t,l} K_W^{l\top}}{\sqrt{d}}\right)V_c^l,
    \label{eq:cross_guidance}
\end{equation}
where $d$ is the key dimension and $V_c^l$ follows the multi-subject ToRA rule in \cref{eq:multi_tora}.
With identical text conditioning and frozen key projections, cross-attention keys are already shared between streams; replacing the cross-attention queries changes where personalized values are read.
The self-attention queries and keys depend on each stream's evolving video latent and therefore generally differ.
We empirically compare different copy strategies in \cref{tab:guidance_methods}.

\myheading{Why copy $Q,K$?}
The queries and keys determine the attention weights, whereas values carry the content being aggregated.
ToRA personalizes the cross-attention values, so copying generic values can suppress the subject information that adaptation learned.
The ablation is consistent with this mechanism: $K,V$ copying yields higher TA (0.714) but lower IP (0.401) than $Q,K$ copying (TA 0.689, IP 0.469).
These results support a design trade-off, rather than proving that attention maps alone determine motion.

We apply the dual-stream design only during the first $T_W$ denoising timesteps and switch back to the single main stream for the remaining steps.
Applying $Q/K$ copying throughout all sampling steps can over-constrain the personalized stream toward the generic base-model appearance.
We ablate $T_W$ and alternative $Q/K/V$ replacement strategies in \cref{subsec:ablation}.

\subsection{Multi-subject DreamBooth Benchmark}
\label{subsec:multidb}

\begin{figure}[tbp]
    \centering
    \includegraphics[width=\linewidth]{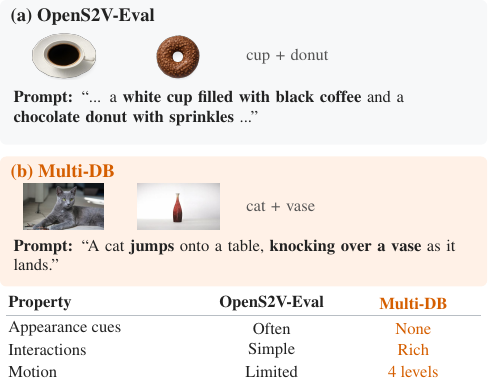}
    \caption{\textbf{What \MultiDB adds.} Appearance-free prompts test reference identity; richer interactions and four motion-difficulty levels stress composition and dynamics. The examples and table compare prompt design with OpenS2V-Eval.}
    \label{fig:dataset}
\end{figure}

Several approaches~\cite{huang2025videomage,chen2024disenstudio} evaluate with private prompts built from image personalization datasets~\cite{ruiz2023dreambooth,kumari2023multi}.
OpenS2V~\cite{yuan2025opens2v} provides an official benchmark for subject-driven video generation, but its multi-subject prompts are simple in motion and interaction, many of them describe the subjects' appearance in detail, which defeats the purpose of personalization, and they do not consistently bind an explicit subject token to each reference image.
Our prompts instead focus on motion and interaction, avoid appearance descriptions, and bind subject tokens to reference images explicitly, as illustrated in \cref{fig:dataset}.

To this end, we introduce \textbf{\MultiDB}, a benchmark for the \textit{multi-subject} setting.
It uses 30 subjects from DreamBench~\cite{ruiz2023dreambooth} covering 15 classes of living and non-living entities, and 122 prompts with two and three subjects, categorized into four motion-difficulty levels.
OpenS2V's evaluated multi-subject subset contains 68 subjects and 32 prompts (24 two-subject and 8 three-subject prompts).
Initial interaction prompts were generated with ChatGPT before any tuning of our method, then manually revised and verified; the instruction, subject statistics, and verification protocol are in the supplement.
\MultiDB is a targeted stress test for three weaknesses of existing benchmarks, namely appearance leakage in prompts, simple interactions and limited motion diversity, and is not meant to replace OpenS2V, which has more subjects but fewer and simpler prompts; we therefore report both.

\section{Experiments}
\label{sec:experiments}

\myheading{Datasets and setting.}
We evaluate on \MultiDB (30 subjects; 76 two-subject and 46 three-subject prompts) and the multi-subject subset of OpenS2V-Eval~\cite{yuan2025opens2v} (68 subjects; 32 prompts).
The main evaluation concerns non-human subjects and uses a single reference image per subject.
\MultiDB tests appearance-free prompts and varied interactions; OpenS2V supplies a complementary public evaluation.

\myheading{Evaluation metrics and motion caveats.}
Following OpenS2V~\cite{yuan2025opens2v}, we report Identity Preservation (IP) and Textual Alignment (TA) using GME~\cite{zhang2024gme,wang2024qwen2}, and Visual Quality (VQ) using an LAION aesthetic predictor~\cite{schuhmann2022improved}.
We also report Motion Smoothness (MS) and Motion Amplitude (MA) in the benchmark and ablation tables.
\textbf{Neither MS nor MA is a reliable optimization target or a test of motion realism.}
MS can reward nearly static scenes, while optical-flow amplitude can increase because of flicker or structural artifacts rather than coherent movement.
Consequently, lower MS or higher MA alone does not establish a motion improvement, and we use no thresholds to declare motion success.

\myheading{Implementation details.}
We use Wan2.1-T2V~\cite{wan2025wan}, 750 adaptation steps per subject, batch size 1, rank 16, AdamW~\cite{loshchilov2017decoupled}, and learning rate $10^{-5}$ with the loss in \cref{eq:flow_matching}.
Inference uses 50 sampling steps, CFG 5.0, $832\times480$ resolution, 81 frames, and $T_W=20$.
We evaluate 1.3B and 14B backbones; ablation sweeps use the 1.3B model and a 20-prompt \MultiDB validation set.
Auxiliary-study settings are reported separately in the supplement.

\myheading{Baselines and comparison scope.}
We compare with training-based Phantom~\cite{liu2025phantom}, VACE~\cite{jiang2025vace}, Skyreels-A2~\cite{fei2025skyreels}, and MAGREF~\cite{deng2025magref}, and with Wan2.1 adaptations of Textual Inversion (TI)~\cite{gal2022image} and DreamBooth~\cite{ruiz2023dreambooth}.

\begin{table*}[tp]
\centering
\small
\renewcommand{\arraystretch}{1.12}
\begin{minipage}[t]{.47\textwidth}
\centering
\setlength{\tabcolsep}{2.5pt}
\caption{\textbf{Results on \MultiDB.} RD denotes extra reference-conditioned video training. MS/MA are descriptive; higher TA/IP are better. Bold marks the best personalized method per model size, excluding the class-only base ($\dagger$).}
\label{tab:quantitative_1}
\begin{tabular}{@{}llccccc@{}}
\toprule
Size & Method & RD & MS & MA & TA$\uparrow$ & IP$\uparrow$ \\
\midrule
\multirow{6}{*}{1.3B}
& Wan2.1 (base)$\dagger$ & & 0.896 & 0.112 & 0.691 & 0.371 \\
& Phantom & \checkmark & 0.911 & 0.089 & 0.669 & \textbf{0.510} \\
& VACE & \checkmark & 0.919 & 0.128 & 0.687 & 0.437 \\
& Textual Inv. & & 0.891 & 0.162 & 0.392 & 0.347 \\
& DreamBooth & & 0.942 & 0.072 & 0.652 & 0.422 \\
& \textbf{Ours} & & 0.908 & 0.125 & \textbf{0.688} & 0.467 \\
\midrule
\multirow{4}{*}{14B}
& VACE & \checkmark & 0.876 & 0.198 & 0.607 & \textbf{0.528} \\
& MAGREF & \checkmark & 0.860 & 0.138 & 0.679 & 0.504 \\
& Textual Inv. & & 0.966 & 0.061 & 0.412 & 0.365 \\
& \textbf{Ours} & & 0.957 & 0.122 & \textbf{0.721} & 0.472 \\
\bottomrule
\end{tabular}
\end{minipage}\hfill
\begin{minipage}[t]{.50\textwidth}
\centering
\setlength{\tabcolsep}{2.5pt}
\caption{\textbf{Results on OpenS2V-Eval.} RD follows \cref{tab:quantitative_1}. MS/MA are descriptive; higher VQ/TA/IP are better. Bold marks the best method per model size.}
\label{tab:quantitative_2}
\vspace{\baselineskip}
\begin{tabular}{@{}llcccccc@{}}
\toprule
Size & Method & RD & VQ$\uparrow$ & MS & MA & TA$\uparrow$ & IP$\uparrow$ \\
\midrule
\multirow{3}{*}{1.3B}
& Phantom & \checkmark & 0.453 & 0.981 & 0.107 & 0.694 & 0.484 \\
& VACE & \checkmark & \textbf{0.455} & 0.992 & 0.105 & \textbf{0.712} & 0.476 \\
& \textbf{Ours} & & 0.369 & 0.958 & 0.109 & 0.637 & \textbf{0.502} \\
\midrule
\multirow{5}{*}{14B}
& Phantom & \checkmark & \textbf{0.444} & 0.982 & 0.218 & 0.694 & 0.496 \\
& VACE & \checkmark & 0.431 & 0.967 & 0.099 & 0.695 & 0.468 \\
& Skyreels-A2 & \checkmark & 0.396 & 0.946 & 0.142 & 0.681 & 0.527 \\
& MAGREF & \checkmark & 0.441 & 0.949 & 0.073 & \textbf{0.709} & \textbf{0.567} \\
& \textbf{Ours} & & 0.387 & 0.974 & 0.126 & 0.696 & 0.482 \\
\bottomrule
\end{tabular}
\end{minipage}
\end{table*}

\subsection{Comparison with State-of-the-Art}
\label{subsec:compare_sota}

\myheading{Quantitative results.}
\Cref{tab:quantitative_1,tab:quantitative_2} summarize results on \MultiDB and OpenS2V-Eval~\cite{yuan2025opens2v}.
On \MultiDB, Textual Inversion has low IP and TA because it fails to capture detailed appearance, while DreamBooth can suffer from subject mixing and static-reference overfitting when independently learned adapters are merged.
Our method balances identity preservation and prompt alignment while composing independently learned subjects.
Training-based methods reach higher absolute identity scores in several settings, and we do not claim otherwise.
On OpenS2V at 14B, MAGREF (IP 0.567) and Skyreels-A2 (IP 0.527) are ahead of \Approach (IP 0.482); at 1.3B, Phantom leads on \MultiDB IP (0.510 vs.\ 0.467).
Within the evaluated optimization-based baselines, the gap in identity and textual alignment is consistently in our favour: on \MultiDB at 1.3B, \Approach improves IP from 0.347 (TI) and 0.422 (DreamBooth) to 0.467, and TA from 0.392 and 0.652 to 0.688.
Among the evaluated personalized methods, \Approach also achieves the best TA on \MultiDB at both model sizes and the best IP at 1.3B on OpenS2V-Eval, without large-scale reference-video training.

\myheading{Computational cost.}
The T2V-guided stream shares one copy of the frozen Wan2.1 weights with the main stream, and ToRA is a per-token low-rank branch used only in the main stream.
Sharing weights avoids doubling the model parameters, but guidance still incurs additional model evaluations, latent state, and activations during the first $T_W=20$ of 50 steps.
``Lightweight'' therefore refers to the per-subject adaptation cost, about 4 minutes and 5.6\,MB without a reference-video dataset, rather than per-video latency.
The supplement reports measured auxiliary inference costs and their settings.

\myheading{Motion-level breakdown.}
\Cref{tab:motion_level} reports the breakdown over the four \MultiDB motion levels.
As prompts become more dynamic, MS decreases and MA increases; TA also decreases, showing that complex subject interactions remain challenging.
These motion statistics are descriptive, as discussed above.

\begin{figure*}[tp]
\centering
\includegraphics{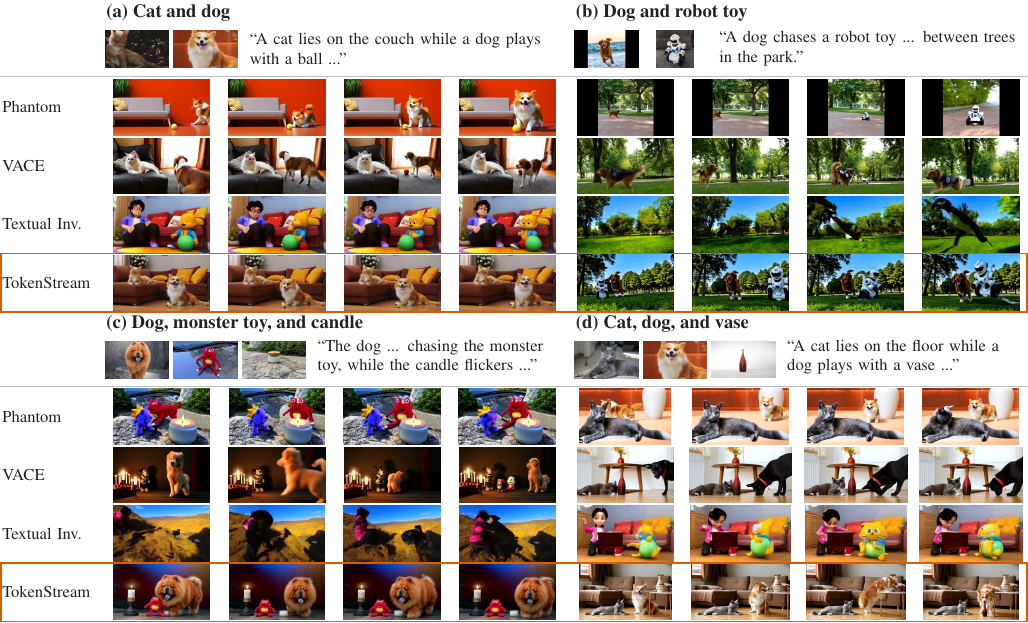}
\caption{\textbf{Qualitative comparisons on \MultiDB.} Selected examples illustrate subject binding and prompt-conditioned composition. They do not establish reliable articulated motion or success on every scene; omission, identity mixing, and same-class failures are shown in the supplement.}
\label{fig:qualitative}
\end{figure*}

\myheading{Qualitative Results.}
\Cref{fig:qualitative} compares methods in multi-object scenarios.
In the examples shown, \Approach separates subjects clearly, whereas Phantom and VACE sometimes omit a requested subject and Textual Inversion fails to capture the subject.

\myheading{User Study.}
We further conduct a user study with 10 participants over four criteria: visual quality (VQ), motion quality (MQ), textual alignment (TA), and identity preservation (IP).
Participants viewed anonymized and shuffled outputs and selected the preferred result under each criterion.
Users preferred \Approach across all criteria (\cref{tab:userstudy}).

\subsection{Ablation Studies}
\label{subsec:ablation}

\begin{table*}[tp]
\centering
\small
\setlength{\tabcolsep}{4pt}
\renewcommand{\arraystretch}{1.05}
\caption{\textbf{Ablation studies.} Each block is an independent sweep. MS/MA are descriptive; higher TA/IP are better. Without ToRA, TI is used.}
\label{tab:ablation_summary}\label{tab:ToRA_WAN}\label{tab:ablation_KV_ToRA}\label{tab:guidance_methods}\label{tab:ablation_selfcross_DSG}\label{tab:ablation_refs}\label{tab:ablation_rank}
\begin{minipage}[t]{.48\textwidth}
\vspace{0pt}
\centering
\begin{tabular}{@{}lrrrr@{}}
\toprule
Setting & MS & MA & TA$\uparrow$ & IP$\uparrow$ \\
\midrule
\multicolumn{5}{@{}l}{\textbf{Components}} \\
TI, no guidance & 0.996 & 0.039 & 0.664 & 0.390 \\
ToRA, no guidance & 0.978 & 0.022 & 0.641 & 0.420 \\
TI + guidance & 0.898 & 0.185 & 0.455 & 0.446 \\
ToRA + guidance & 0.878 & 0.199 & \textbf{0.689} & \textbf{0.469} \\
\midrule
\multicolumn{5}{@{}l}{\textbf{Guided attention layers}} \\
Cross-attention only & 0.900 & 0.204 & \textbf{0.692} & 0.416 \\
Self-attention only & 0.864 & 0.189 & 0.681 & 0.449 \\
Self- and cross-attention & 0.878 & 0.199 & 0.689 & \textbf{0.469} \\
\midrule
\multicolumn{5}{@{}l}{\textbf{Reference images per subject}} \\
1 image & 0.920 & 0.214 & 0.696 & 0.436 \\
2 images & 0.947 & 0.130 & \textbf{0.697} & 0.458 \\
3 images & 0.938 & 0.167 & 0.694 & 0.477 \\
4 images & 0.947 & 0.168 & 0.695 & \textbf{0.515} \\
\bottomrule
\end{tabular}
\end{minipage}
\hfill
\begin{minipage}[t]{.48\textwidth}
\vspace{0pt}
\centering
\begin{tabular}{@{}lrrrr@{}}
\toprule
Setting & MS & MA & TA$\uparrow$ & IP$\uparrow$ \\
\midrule
\multicolumn{5}{@{}l}{\textbf{Adapted projections}} \\
To-V only & 0.988 & 0.041 & 0.607 & \textbf{0.439} \\
To-K and To-V & 0.978 & 0.022 & \textbf{0.641} & 0.420 \\
\midrule
\multicolumn{5}{@{}l}{\textbf{Copied attention features}} \\
None & 0.978 & 0.022 & 0.641 & 0.420 \\
$Q,V$ & 0.862 & 0.212 & 0.700 & 0.453 \\
$K,V$ & 0.899 & 0.215 & \textbf{0.714} & 0.401 \\
$Q,K$ & 0.878 & 0.199 & 0.689 & \textbf{0.469} \\
\midrule
\multicolumn{5}{@{}l}{\textbf{ToRA rank}} \\
16 & 0.879 & 0.188 & 0.695 & 0.463 \\
32 & 0.878 & 0.199 & 0.689 & \textbf{0.469} \\
64 & 0.853 & 0.195 & 0.694 & 0.456 \\
128 & 0.878 & 0.211 & \textbf{0.697} & 0.460 \\
\bottomrule
\end{tabular}
\end{minipage}
\end{table*}

\begin{table}[tb]
\centering
\small
\setlength{\tabcolsep}{5pt}
\renewcommand{\arraystretch}{1.08}
\caption{\textbf{User-study preferences (\%).} Ten participants.}
\label{tab:userstudy}
\begin{tabular}{@{}lrrrr@{}}
        \toprule
        \textbf{Method} & \textbf{VQ} & \textbf{MQ} & \textbf{TA} & \textbf{IP} \\
        \midrule
        Textual Inv. & 0.8 & 0.8 & 0.8 & 0.0 \\
        DreamBooth & 5.7 & 10.6 & 10.6 & 4.1 \\
        Phantom & 10.6 & 11.4 & 10.6 & 12.2 \\
        VACE & 12.2 & 13.0 & 10.6 & 8.1 \\
        \textbf{\Approach} & \textbf{70.7} & \textbf{64.2} & \textbf{65.9} & \textbf{75.6} \\
        \bottomrule
    \end{tabular}

\vspace{8pt}
\caption{\textbf{Results across \MultiDB motion levels.}}
\label{tab:motion_level}
\begin{tabular}{@{}ccccc@{}}
    \toprule
    \textbf{Level} & \textbf{MS} & \textbf{MA} & \textbf{TA}$\uparrow$ & \textbf{IP}$\uparrow$ \\
    \midrule
    1 & 0.975 & 0.012 & \textbf{0.755} & 0.410 \\
    2 & 0.958 & 0.043 & 0.689 & 0.415 \\
    3 & 0.864 & 0.185 & 0.677 & 0.401 \\
    4 & 0.822 & 0.218 & 0.652 & \textbf{0.535} \\
    \bottomrule
    \end{tabular}
\end{table}

\myheading{What each component contributes.}
In the component sweep of \cref{tab:ablation_summary}, replacing TI with ToRA without guidance raises IP from 0.390 to 0.420, while TA falls from 0.664 to 0.641.
More adaptation capacity therefore improves appearance but does not by itself retain prompt-conditioned dynamics.
Adding guidance to ToRA raises both TA (0.641 to 0.689) and IP (0.420 to 0.469).
Guidance with TI instead raises IP but sharply lowers TA to 0.455, indicating that borrowing attention structure is insufficient without a suitable personalized value path.

\myheading{Why adapt values and retain personalized content.}
The projection sweep favors To-V alone for IP (0.439 versus 0.420), whereas adding To-K adaptation improves TA (0.641 versus 0.607).
Restricting adaptation to values leaves the base token-to-layout mechanism intact and offers a specific identity-oriented design choice.
In the separate copy sweep, $K,V$ copying gives higher TA but lower IP than $Q,K$ copying.
That pattern is consistent with generic values weakening personalized appearance, and supports retaining the learned value path rather than copying all available features.
Guiding both self- and cross-attention gives higher IP than either alone (0.469 versus 0.449/0.416), while cross-attention-only guidance gives slightly higher TA.
The choice therefore reflects a trade-off, not dominance on every score.

\begin{figure}[t]
\centering
\includegraphics[width=\linewidth]{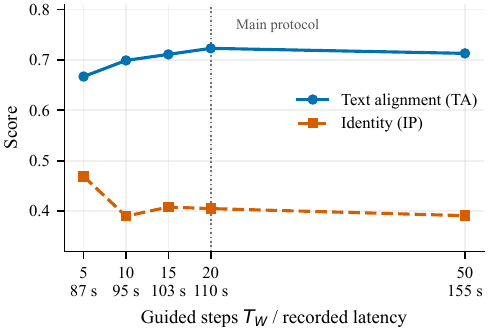}
\caption{\textbf{Guidance-window trade-off.} Textual alignment improves up to 20 guided steps, then declines. Identity preservation decreases overall with longer guidance, while sampling latency increases.}
\label{fig:guidance_window}\label{tab:ablation_timestepDSG}
\end{figure}

\myheading{Guidance-window length.}
\Cref{fig:guidance_window} shows the identity--alignment trade-off.
Extending guidance from 5 to 20 steps raises TA from 0.667 to 0.723 and latency from 87 to 110\,s, while IP changes from 0.469 to 0.405.
We use $T_W=20$ for alignment.
Guiding all 50 steps is slower and lowers both scores relative to 20 steps.

\myheading{Reference count and adaptation rank.}
The sweeps in \cref{tab:ablation_summary} show that increasing reference images from one to four raises IP from 0.436 to 0.515 while TA remains within 0.694--0.697.
Additional views can thus help appearance learning, but require more user input.
Larger LoRA rank does not consistently improve IP.

\myheading{Scaling and remaining limitations.}
The supplementary study composes $N\in\{2,3,4,6\}$ adapters, with ten clips per setting.
IP decreases from 0.487 to 0.218 and detector-based missing subject--frame pairs rise from 0.189 to 0.560 between $N=2$ and $N=6$ despite nearly flat measured resource use.
Scene complexity and subject combinations also vary, so these results do not isolate the cause of failures.
Same-class fusion and human-identity failures are reported in the supplement.
These results distinguish computational scalability from generation quality.
Independent adapters remain reusable across prompts and subject combinations without joint optimization or weight averaging, even as identity preservation degrades and omissions increase when more subjects are composed.

\par
\makeatletter
\if@firstcolumn
  \newpage
  \null
\fi
\newpage
\makeatother
\section{Conclusion}
\label{sec:conclusion}

In this paper, we introduced \Approach, a novel framework for multi-subject personalized video generation from independently adapted subjects.
Our method leverages \textbf{Token-wise LoRA} for flexible multi-subject control and the \textbf{T2V-guided stream} to reuse pretrained attention structure during inference.
We also proposed \textbf{\MultiDB}, a new benchmark featuring more complex motion and interactions to better evaluate multi-subject generation.
Experiments on \MultiDB and OpenS2V show improved identity preservation and textual alignment over the evaluated TI and DreamBooth baselines, with complementary trade-offs against training-based methods.

\Approach still has limitations.
Our evaluation focuses on two to three non-human subjects, primarily from distinct classes; human identity, same-class separation, many-subject composition, and articulated animal motion remain challenging.
Large training-based methods retain advantages on several metrics, and extreme subject-prior geometry mismatch can bias the guided trajectory.

\clearpage
{\small
\bibliographystyle{ieeenat_fullname}
\bibliography{main}
}
\twocolumn[{\centering{\Large\bf Supplementary Material}\par\vspace{1.5em}}]
\appendix
\section{Experimental Details}
\label{sec:supp_exp_details}

We use a single NVIDIA H100 GPU for both training and inference of \Approach.
All experiments are conducted with Wan2.1-T2V~\cite{wan2025wan}.
For each subject, we train ToRA for 750 steps with batch size 1, learning rate $1\times 10^{-5}$, LoRA rank 16, AdamW, and the flow matching loss.
For inference, we use 50 sampling steps, classifier-free guidance scale 5.0, and resolution $832 \times 480$.
Each generated video contains 81 frames. The main protocol uses $T_W=20$.
The auxiliary scaling, prompt-sensitivity, background and portrait studies instead use the recorded 500-step, $10^{-4}$ learning-rate, rank-16 recipe with $T_W=10$; their detection-based scores are separate from the main benchmarks.

\myheading{Prompts used for subject learning.}
Previous optimization-based approaches~\cite{ruiz2023dreambooth,gal2022image} often use simple template prompts such as ``a photo of a dog'' for subject learning.
In our implementation, each reference image is paired with a detailed caption describing the visible subject and its surrounding context.
The caption helps the frozen text encoder align the reference image with the corresponding subject noun while keeping the actual generation prompt free to specify new actions and interactions.
For example, the dog in the teaser is captioned as:
\textit{``A dog resting comfortably on a couch in a cozy indoor setting. The couch is upholstered in dark gray fabric with a textured pattern, and there is a decorative pillow featuring a geometric design in dark green with white lines. In the background, a window with horizontal blinds allows some light to filter into the room, and dark curtains hang beside it. The overall ambiance suggests a relaxed and homely environment.''}

\myheading{Textual Inversion on Wan2.1.}
We adapt Textual Inversion to Wan2.1 by optimizing the textual embedding of the umT5~\cite{chung2023unimax} encoder instead of the CLIP~\cite{radford2021learning} text encoder used in the original image-generation setting.
We train each subject with learning rate $1\times 10^{-5}$ for 1000 steps and batch size 1.

\myheading{Runtime.}
We report the total running time of \Approach in \cref{tab:supp_runtime}, including ToRA optimization and T2V-guided inference.
The optimization time is for a single subject, while nearly flat inference time is observed only for the evaluated subject counts.
The numbers are measured with Wan2.1-T2V-1.3B on a single H100; the 110-second inference time is from the $T_W=20$ window sweep, separate from the auxiliary timings below.

\begin{table}[t]
    \centering
    \footnotesize
    \caption{Runtime of ToRA optimization and T2V-guided inference on a single H100.}
    \label{tab:supp_runtime}
    \begin{tabular}{@{}lc@{}}
        \toprule
        \textbf{Phase} & \textbf{Time} \\
        \midrule
        Optimization (ToRA, per subject) & $\sim$4 min \\
        Inference (T2V-guided stream) & $\sim$1 min 50 s \\
        \bottomrule
    \end{tabular}
\end{table}

\myheading{User study.}
The main-paper user study uses 10 participants. Original vote records and comparison assignments are unavailable, so the reported preferences are descriptive and do not support a valid uncertainty estimate.

\section{Success and Failure Cases Under Shape Mismatch}
\label{sec:supp_success_failure}

\begin{figure*}[t]
    \centering
    \includegraphics[width=\textwidth]{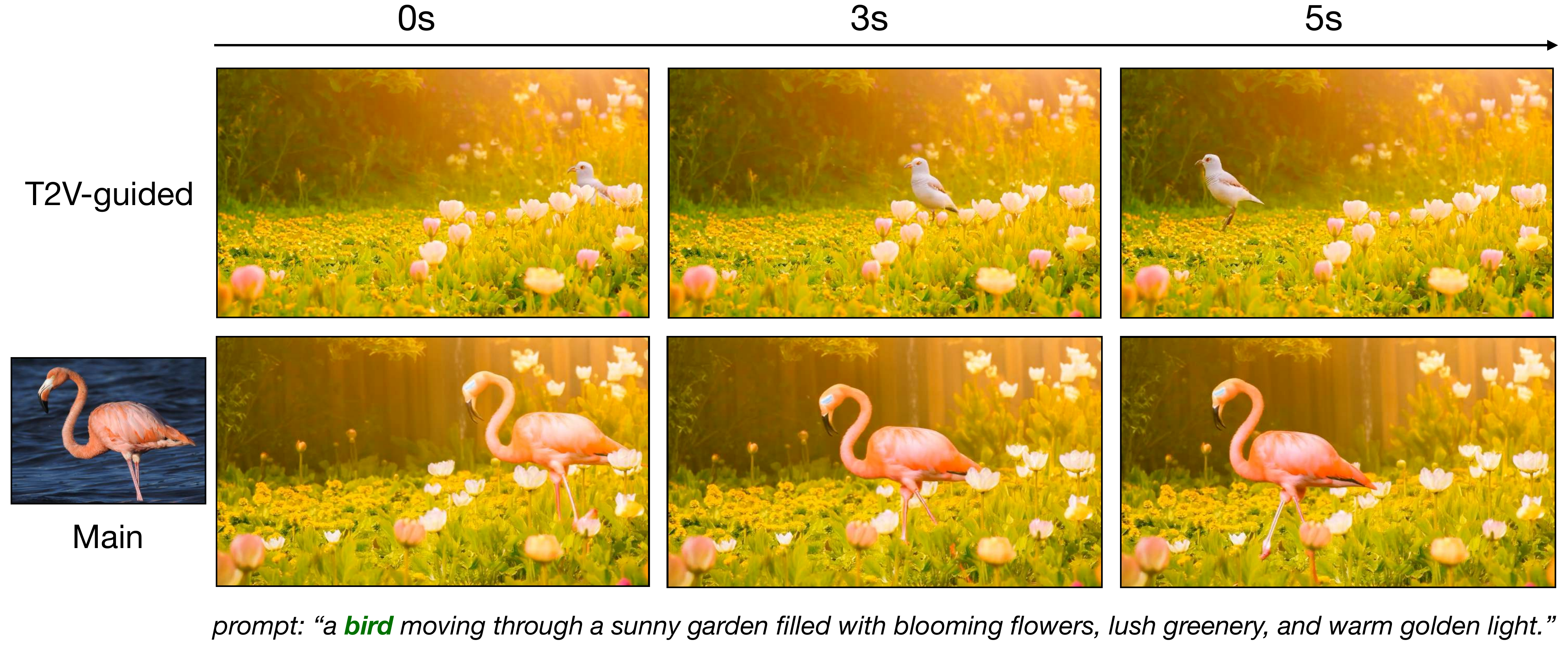}
    \caption{Success case under moderate shape mismatch.
    \Approach preserves the long legs of a Flamingo despite mismatch with the generic ``bird'' prior.}
    \label{fig:supp_success}
\end{figure*}

\begin{figure*}[t]
    \centering
    \includegraphics[width=\textwidth]{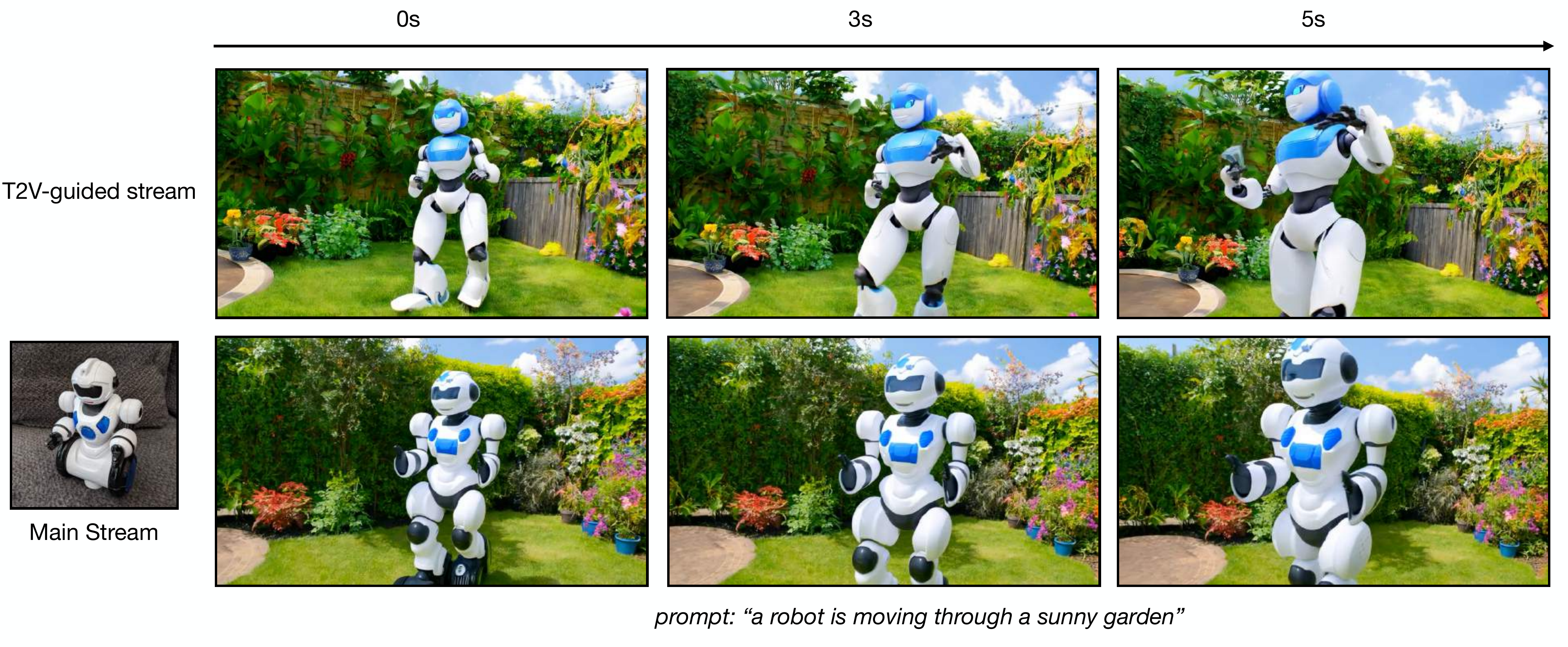}
    \caption{Failure cases under stronger prior mismatch.
    When the subject geometry or expected dynamics strongly differs from the generic T2V prior, the guided stream can bias the coarse layout or motion.
    These examples illustrate the limitation discussed in the main paper.}
    \label{fig:supp_failure}
\end{figure*}

The T2V-guided stream transfers coarse structure and motion from the base Wan2.1 stream by copying $Q/K$ features during the first $T_W$ denoising steps.
This guidance acts as an early scaffold rather than a hard constraint because the personalized stream still carries subject-specific values and residual updates.
As a result, \Approach can preserve moderate subject-shape deviations (shown in ~\cref{fig:supp_success}), but it can still inherit incorrect coarse motion or geometry when the reference subject deviates significantly from the generic prior, as illustrated in ~\cref{fig:supp_failure}.

\section{\MultiDB benchmark}
\label{sec:supp_multidb}

\MultiDB is designed as a targeted stress test for multi-subject personalization rather than a comprehensive benchmark covering all possible subject combinations. It specifically addresses three limitations of existing benchmarks: appearance leakage from textual descriptions, weak inter-subject interactions, and insufficient motion diversity.

\myheading{Subject and prompt statistics.}
\MultiDB uses 30 subjects from DreamBench~\cite{ruiz2023dreambooth}, covering 15 living and non-living classes.
It contains 122 prompts, including 76 two-subject prompts and 46 three-subject prompts.
The prompts are assigned to four motion-difficulty levels, ranging from minimal motion to highly complex dynamics.

\myheading{Prompt construction.}
We first generated candidate interaction prompts before method-specific tuning.
To construct meaningful subject combinations, we randomly sampled and shuffled subjects from the DreamBench subject set, then prompted GPT to generate a natural video-generation prompt describing interactions among the selected subjects.
The generated prompts were subsequently manually revised to remove explicit appearance descriptions and to ensure that each subject noun is unambiguously associated with its reference image.
For example, a prompt may specify that a ``dog'' plays with a ``toy car'', but should not describe the dog's color, breed, fur texture, or any other identity-specific attributes, which should instead be inferred from the reference image.
For reproducibility, we include the instruction used for prompt generation below.

\begin{quote}\small
\textbf{GPT prompt.}
You are given a set of subject names: \texttt{\{SUBJECTS\}}.
Generate a single natural video-generation prompt that includes all of the given subjects in one coherent scene.

\textbf{Requirements.}
\begin{itemize}[leftmargin=*,noitemsep,topsep=2pt]
    \item Every subject must appear exactly once.
    \item All subjects should participate in meaningful interactions.
    \item Do not describe any appearance attributes, including color, texture, breed, clothing, age, size, identity, or style.
    \item Refer to each subject only by its noun, e.g., ``dog'', ``toy car'', or ``backpack''.
    \item Focus on actions, interactions, scene layout, camera motion, and subject motion.
    \item Produce a single concise paragraph suitable for text-to-video generation.
\end{itemize}
\end{quote}

\myheading{Motion-level annotation.}
The four levels describe increasing motion and interaction difficulty.
Level 1 contains mostly static or low-motion scenes, while Level 4 contains stronger subject movement, interaction, or changing spatial relations.
Each prompt was assigned a level by two annotators independently, using the number of moving subjects, the presence of subject-subject interaction, and whether the spatial relation between subjects changes over the clip; disagreements were resolved by discussion.

\myheading{On using an LLM to draft prompts.}
Prompts were drafted by GPT and then revised by hand, and it is fair to ask what bias that introduces.
Three effects are worth naming.
First, LLM-drafted prompts favour a recognizable register: common verbs (``chases'', ``rolls'', ``jumps''), everyday indoor and park settings, and a tendency toward benign, well-formed scenes. \MultiDB therefore under-represents unusual or adversarial compositions.
Second, an LLM asked to make subjects interact will reliably produce interaction, which is what we wanted, but it also means interaction is distributed fairly evenly rather than sampled from any natural distribution of user requests.
Third, and most relevant to fairness between methods, an LLM prompted to avoid appearance words does not always succeed; this is why every prompt was manually revised to remove residual appearance cues and to check that each subject noun maps unambiguously to one reference image.
The manual revision stage is what makes the benchmark usable for personalization, and it was carried out before any tuning of our method, which reduces method-specific tuning but does not establish the absence of benchmark bias.
We consider the residual register bias a genuine limitation, shared with any LLM-generated benchmark, and one reason we report OpenS2V results alongside \MultiDB rather than in place of them.

\section{Additional Ablation on Token-wise LoRA Rank}
\label{sec:supp_rank}

We further investigate the impact of LoRA rank in ToRA.
As shown in main-paper Table~3, rank 16 and rank 32 provide a good balance between identity preservation and textual alignment.
Very large ranks do not consistently improve identity preservation, suggesting that additional capacity can overfit the static reference or interfere with prompt following.

\section{Impact of Timesteps Used for the T2V-Guided Stream}
\label{sec:supp_tw}

We also provide qualitative results for the number of timesteps used in the T2V-guided stream.
Applying guidance throughout the full 50-step denoising process can degrade subject identity preservation, often producing a video closer to the generic Wan2.1 output.
Reducing $T_W$ to 40 or 30 steps may still blend the personalized subject with the generic subject from the base stream.
Since our goal is to retain only the structural layout and motion from the base stream, we use $T_W=20$ for textual alignment; the reported window sweep has higher IP at $T_W=5$.
The qualitative comparison is provided in the supplementary video file \texttt{qualitative/ablate\_timesteps.html}.

\section{Additional Qualitative Results and Supplementary Videos}
\label{sec:supp_qualitative}

We provide additional qualitative comparisons on \MultiDB.
Phantom sometimes omits one of the requested subjects and can exhibit copy-and-paste artifacts where the reference image background is reproduced in the generated video.
VACE often follows the prompt but has weaker identity preservation.
Textual Inversion struggles to capture detailed subject identity, while DreamBooth with averaged LoRA weights can introduce subject-mixing artifacts.
In contrast, \Approach generally preserves separate subject identities while maintaining prompt alignment.
For a detailed comparison, we recommend viewing the supplementary videos in the static file \texttt{qualitative/multi\_db.html}.

\section{Computational Cost}
\label{sec:supp_cost}

\Cref{tab:cost} compares training requirements, per-subject adaptation cost, storage, inference latency and peak memory.

Three points are worth making explicit.
First, the T2V-guided stream does not instantiate a second model. Both streams evaluate the \emph{same} frozen Wan2.1 weights, and ToRA is a low-rank branch on the To-V projection that is active only in the main stream, so the additional state is one extra latent together with its activations. Guidance still adds base-model evaluations and activation memory during the first $T_W$ sampling steps; sharing weights does not eliminate inference overhead.
Second, Textual Inversion and DreamBooth modify only weights or embeddings and leave the sampling loop untouched, so they avoid the extra guided evaluations. Matched comparative inference costs were not measured; DreamBooth additionally requires a merging step whenever a new combination of subjects is requested.
Third, the training-based baselines are feed-forward for a new subject, but that convenience is purchased with a curated corpus of reference-video pairs and a full training run, neither of which appears in a per-video latency measurement. The comparison is therefore between a very large one-time cost and a small recurring one, and which is preferable depends on how often a subject is reused.

We measured latency and peak allocated memory for \Approach on one H100 under the auxiliary $T_W=10$ configuration, averaged over three repetitions of a fixed prompt/seed. Recovered auxiliary code captures unadapted attention on the personalized latent; these historical timings do not measure an independently evolving base trajectory. The Phantom and VACE entries are \emph{estimates} rather than measurements: both are single-pass Wan2.1 samplers that add a small number of reference-conditioning tokens, so their per-video cost should sit slightly above the single-stream baseline. We mark them accordingly rather than presenting borrowed numbers as our own.

\begin{table}[tbp]
    \centering
    \footnotesize
    \setlength{\tabcolsep}{2.45pt}
    \renewcommand{\arraystretch}{1.08}
    \caption{Cost comparison on one H100 with Wan2.1-T2V-1.3B, 50 steps, $832\times480$, 81 frames ($T_W=10$ for \Approach). Entries marked $\approx$ are estimated, not measured; the \Approach row is the mean of three timed runs.}
    \label{tab:cost}
    \begin{tabular}{@{}lccccc@{}}
        \toprule
        \textbf{Method} & \makecell{\textbf{Ref.-video}\\\textbf{dataset}} & \makecell{\textbf{Cost per}\\\textbf{new subj.}} & \makecell{\textbf{Store}\\\textbf{/subj.}} & \textbf{Latency} & \makecell{\textbf{Peak}\\\textbf{mem.}} \\
        \midrule
        Phantom & required & feed-fwd & --- & $\approx$2m10s & $\approx$24\,GiB \\
        VACE & required & feed-fwd & --- & $\approx$2m15s & $\approx$24\,GiB \\
        \midrule
        Textual Inv. & none & $\sim$4\,min & 8\,KB & $\approx$2m10s & $\approx$24\,GiB \\
        DreamBooth & none & $\sim$4\,min\,+\,merge & 38\,MB & $\approx$2m10s & $\approx$24\,GiB \\
        \textbf{Ours} & none & $\sim$4\,min & 5.6\,MB & \textbf{2m11s} & \textbf{23.74\,GiB} \\
        \bottomrule
    \end{tabular}
\end{table}

\section{Scaling to More Subjects}
\label{sec:supp_scaling}

\Cref{tab:scaling} composes $N\in\{2,3,4,6\}$ independently trained ToRAs, ten videos per row. The behaviour separates into two parts.

\myheading{Cost composes.} Latency stays within 129--132\,s and peak memory within 23.74--23.75\,GiB across the whole range, nearly flat over the evaluated counts. Adapter storage grows by about 5.6\,MB per subject, and token-projection work also grows with routed spans.

\myheading{Quality does not.} Identity preservation falls from 0.487 at $N\!=\!2$ to 0.218 at $N\!=\!6$, and the mean missing subject--frame rate rises from 0.19 to 0.56. \Cref{fig:supp_scaling} shows both failure modes: subjects go missing, and those that remain fuse with one another. Our benchmarks target two to three non-human, distinct-class subjects, which can also exhibit failures; many-subject composition remains future work.

\begin{figure}[tbp]
    \centering
    \includegraphics[width=\linewidth]{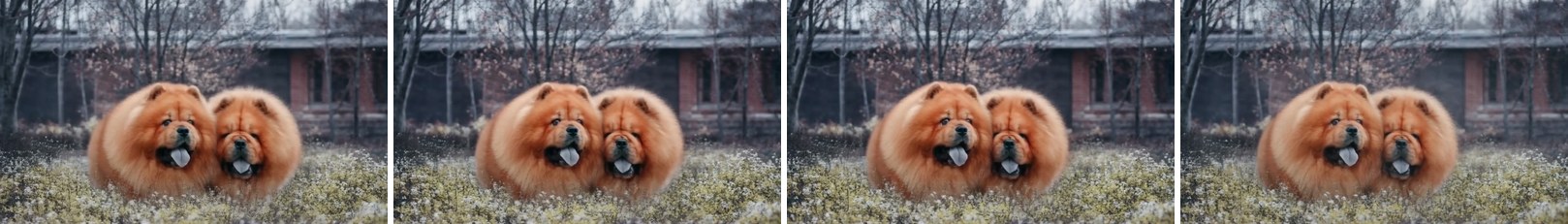}
    \caption{Two subjects of the same class (``\emph{dog1} and \emph{dog2} play together in a yard''). The two instances are rendered as one fused animal rather than two identities. We do not claim composability in this setting.}
    \label{fig:supp_sameclass}
\end{figure}

\myheading{Same-class subjects.} Composing two subjects that share a class noun does not yield two distinct identities (\cref{fig:supp_sameclass}), and rewriting the prompt to give each a distinguishing role phrase lowers identity preservation from 0.515 to 0.431. We do not claim composability for same-class subjects.

These studies use the repository default inference configuration and a detection-based identity score that penalises omission directly, so their values are internally comparable but not comparable to Tabs.~1 and~2 of the main paper.
``Missing'' averages undetected requested subject--frame pairs over 32 uniformly sampled frames per clip.\footnote{The recorded detector is Grounding DINO Tiny (\texttt{IDEA-Research/}\allowbreak\texttt{grounding-dino-tiny}). Recovered code defaults to box/text thresholds 0.25/0.20 and rejects boxes smaller than eight pixels; historical invocation overrides remain unverified.}
Detector error bounds were not measured: missed detections can inflate the rate, while fused detections can reduce it. Changing subject count also changes scene complexity, so the omission trend does not isolate token competition from scene crowding or base-model limitations.

\begin{table}[tbp]
    \centering
    \footnotesize
    \setlength{\tabcolsep}{3.4pt}
    \renewcommand{\arraystretch}{1.08}
    \caption{Composing $N$ independently trained ToRAs, and the same-class case. ``Missing'' is the mean fraction of undetected requested subject--frame pairs.}
    \label{tab:scaling}
    \begin{tabular}{@{}lccccc@{}}
        \toprule
        \textbf{Setting} & \textbf{TA}$\uparrow$ & \textbf{IP}$\uparrow$ & \textbf{Missing}$\downarrow$ & \textbf{Latency} & \textbf{Peak mem.} \\
        \midrule
        $N=2$ & 0.620 & 0.487 & 0.189 & 130.95\,s & 23.742\,GiB \\
        $N=3$ & 0.611 & 0.328 & 0.405 & 128.81\,s & 23.745\,GiB \\
        $N=4$ & 0.494 & 0.282 & 0.425 & 131.52\,s & 23.747\,GiB \\
        $N=6$ & 0.574 & 0.218 & 0.560 & 132.33\,s & 23.753\,GiB \\
        \midrule
        Same class, $N=2$ & 0.675 & 0.515 & 0.150 & 128.69\,s & 23.742\,GiB \\
        \; with role phrases & 0.652 & 0.431 & 0.234 & 130.20\,s & 23.742\,GiB \\
        \bottomrule
    \end{tabular}
\end{table}

\begin{figure*}[tbp]
    \centering\footnotesize
    \setlength{\tabcolsep}{2pt}\renewcommand{\arraystretch}{1.1}
    \begin{tabular}{@{}l@{\hspace{3pt}}c@{}}
      $N\!=\!4$ & \raisebox{-0.5\height}{\includegraphics[width=.93\textwidth]{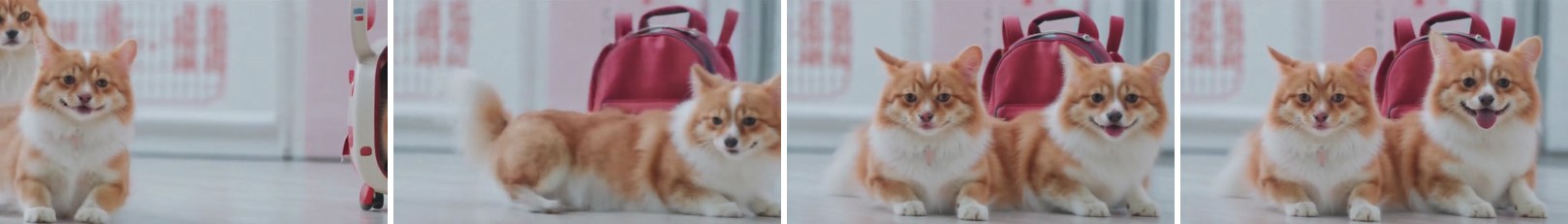}}\\
      $N\!=\!6$ & \raisebox{-0.5\height}{\includegraphics[width=.93\textwidth]{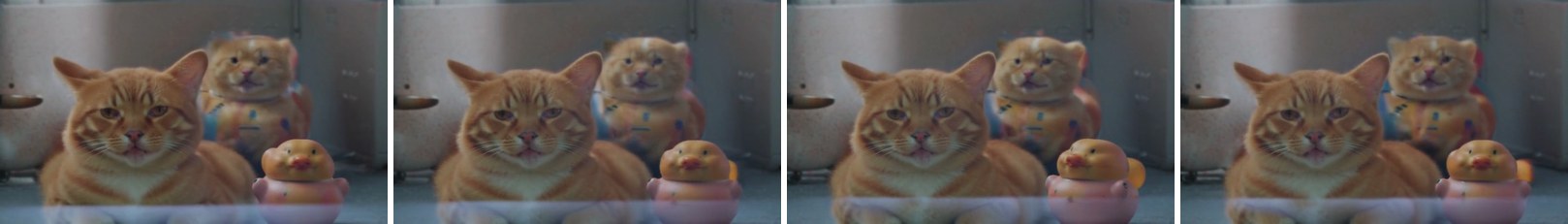}}\\
    \end{tabular}
    \caption{Composing many subjects, four frames from the median-IP clip of each group. Requested subjects are missing and the remaining ones fuse together, while the main benchmarks target two to three non-human subjects.}
    \label{fig:supp_scaling}
\end{figure*}

\section{Isolating the Prompt-Fitting Step}
\label{sec:supp_prompt_fitting}

\MultiDB prompts deliberately contain no appearance words, and the OpenS2V results in the main paper use the official prompts unchanged. It is still fair to ask how much of a method's score comes from the prompt describing the subjects rather than from the method itself.

\Cref{tab:supp_prompt} evaluates each optimization-based method twice on the same two-subject scenes, once with our appearance-free prompts and once with appearance words restored. Restoring them raises textual alignment for every method, and raises identity preservation for Textual Inversion and DreamBooth. These examples indicate that appearance-free prompts are demanding for all three evaluated methods, without establishing the absence of benchmark bias. This is a diagnostic with one video per cell, so we claim no ranking between methods.

\begin{table}[tbp]
    \centering
    \footnotesize
    \setlength{\tabcolsep}{5pt}
    \renewcommand{\arraystretch}{1.06}
    \caption{Prompt-sensitivity diagnostic on two-subject scenes. One video per cell, so this illustrates prompt sensitivity rather than a ranking.}
    \label{tab:supp_prompt}
    \begin{tabular}{@{}llcc@{}}
        \toprule
        \textbf{Method} & \textbf{Prompt} & \textbf{TA}$\uparrow$ & \textbf{IP}$\uparrow$ \\
        \midrule
        \multirow{2}{*}{Textual Inv.}
          & appearance-free & 0.726 & 0.423 \\
          & appearance-revealing & 0.777 & 0.536 \\
        \midrule
        \multirow{2}{*}{DreamBooth}
          & appearance-free & 0.683 & 0.651 \\
          & appearance-revealing & 0.819 & 0.702 \\
        \midrule
        \multirow{2}{*}{\textbf{Ours}}
          & appearance-free & 0.646 & 0.668 \\
          & appearance-revealing & 0.858 & 0.634 \\
        \bottomrule
    \end{tabular}
\end{table}

\section{Background Memorization}
\label{sec:supp_background}

Our reference captions describe the subject and its surroundings, so it is fair to ask whether ToRA stores the context as well as the subject.

We generated 15 videos per method, placing each of three subjects in five scenes incompatible with its reference context. Subjects are localized with Grounding DINO Tiny and refined with SAM ViT-B. Recovered evaluator code replaces masked foreground pixels by RGB $(127,127,127)$ and computes cosine similarity between L2-normalized final-layer DINOv2-base CLS features, averaging eight frames per clip and then clips; historical invocation overrides and model revisions remain unverified. The base model does not receive the reference and provides an empirical comparison level.

\Cref{tab:supp_background} puts \Approach at 0.142, close to the base value of 0.121, while DreamBooth reaches 0.261. This indicates less reference-background similarity than DreamBooth under this diagnostic, rather than proving that ToRA stores no background context. Restricting the comparison to clips in which the subject is detected in every frame leaves the ordering intact (0.163 base, 0.195 ours, 0.295 DreamBooth), although detection and segmentation differences remain possible confounds. Identity preservation was not re-measured on this auxiliary set.

\begin{table}[tbp]
    \centering
    \footnotesize
    \renewcommand{\arraystretch}{1.08}
    \setlength{\tabcolsep}{5pt}
    \caption{Background similarity between generated non-subject regions and the reference background. The base model does not receive the reference and provides an empirical comparison level.}
    \label{tab:supp_background}
    \begin{tabular}{@{}lcc@{}}
        \toprule
        \textbf{Method} & \makecell{\textbf{Bkg.\ sim.\ to ref.}$\downarrow$} & \makecell{\textbf{Subject}\\\textbf{detection rate}} \\
        \midrule
        Wan2.1 (base) & \textbf{0.121} & 0.758 \\
        DreamBooth & 0.261 & 0.867 \\
        \textbf{Ours} & 0.142 & 0.708 \\
        \bottomrule
    \end{tabular}
\end{table}

\section{Multi-Human Identity}
\label{sec:supp_human}

DreamBench, on which \MultiDB is built, contains no human subjects, so our main benchmarks do not evaluate face identity. We trained portrait adapters with the auxiliary recipe above and composed them in two-person prompts.

\textbf{The method does not reliably preserve human identities.} The generated people do not carry the reference identities, so we report no numbers here and state the limitation plainly rather than presenting a favourable subset.

A single-frame reference provides limited viewpoint supervision, and the base prior can influence geometry; these examples do not isolate the cause of identity drift.

Dedicated identity benchmarks such as the subject mode of VideoAlchemist~\cite{chen2024videoalchemy} are the right evaluation for this question; we could not run them because their evaluation code and reference splits were not available to us. Multi-view references and a tighter guidance schedule are the clearest directions for future work.

\end{document}